# AN EFFICIENT FEATURE SELECTION IN CLASSIFICATION OF AUDIO FILES

Jayita Mitra [1] and Diganta Saha [2]

[1] Assistant Professor, Dept. of IT, Camellia Institute of Technology, Kolkata, India
jmitra2007@yahoo.com
[2] Associate Professor, Department of CSE, Jadavpur University, Kolkata, India
neruda0101@yahoo.com

## ABSTRACT

*In this paper we have focused on an efficient feature selection method in classification of audio files. The main objective is feature selection and extraction. We have selected a set of features for further analysis, which represents the elements in feature vector. By extraction method we can compute a numerical representation that can be used to characterize the audio using the existing toolbox. In this study Gain Ratio (GR) is used as a feature selection measure. GR is used to select splitting attribute which will separate the tuples into different classes. The pulse clarity is considered as a subjective measure and it is used to calculate the gain of features of audio files. The splitting criterion is employed in the application to identify the class or the music genre of a specific audio file from testing database. Experimental results indicate that by using GR the application can produce a satisfactory result for music genre classification. After dimensionality reduction best three features have been selected out of various features of audio file and in this technique we will get more than 90% successful classification result.*

## 1. INTRODUCTION

Data mining is the process of analyzing the data and discovering previously unknown pattern from large dataset. The data sources can be any databases, the web, data warehouses, transactional data, data streams, spatial data, or information repositories. The aim of this process is to extract information and summarizing it into an understandable structure for further use. Data mining functionalities include discovering frequent patterns, associations, and correlations; classification and regression; and clustering analysis are found in [1].

Classification is the task of generalizing known structure to apply to the new dataset. Data classification is a two-step process learning or training phase and classification step. In learning phase a classification model is constructed which describes a predetermined set of data classes or concepts. In case of classification the test data are used to estimate the accuracy of the classification rules. The accuracy of a classifier on a test data set is the percentage of the test data set tuples that are correctly classified by the classifier. If the accuracy of the classifier is considered acceptable, the classifier can be used to classify future data set for which the class labels unknown.

Feature selection is the process of selecting a subset of relevant features by eliminating features with no predictive information and selected features are further used in classifier model construction. The data set may contain redundant or irrelevant features. Redundant features are not providing any information in comparing to currently selected features, and irrelevant features provide no useful information in any context. Feature selection techniques are often used in domains where there are many features and comparatively few data points. The basic objectives of this technique are to avoid overfitting, provide faster and cost-effective models and improvement of model performance and efficiency. The usefulness of feature selection is to reduce the noise for improvement of accuracy of classification, interpretable features to identify the function type and dimensionality reduction to improve the computational cost. Feature extraction is a special form of dimensionality reduction in which extracted features are selected such a manner that the feature set will extract relevant information from large data set to achieve the goal using reduced data set.

An attribute selection measure provides rankings for each attribute describing a set of training tuples mentioned in [1]. The attribute which is having maximum value for the measure is chosen as splitting attribute. The splitting criterion indicates the splitting attribute and may also indicate a split-point or a splitting subset. More specifically it selects an attribute by determining the best way to separate or partition the tuples into individual classes.

The paper [2] briefly focused on feature construction, feature ranking, multivariate feature selection, efficient search methods, and feature validity assessment methods. It also described filters that select variables by ranking them with correlation coefficients. The subset selection method is also discussed here. It also includes wrapper methods that assess subsets of variables according to their usefulness to a given predictor.

The paper is organized as follows: Section II starts by describing the literature survey and related work of this paper. Section III includes system design and module description. Feature selection is introduced in Section IV. Section V mainly focused on Gain Ratio. We briefly discuss data analysis and experimental evaluation in Section VI. Finally, we conclude the paper in the last section.

## 2. LITERATURE SURVEY AND RELATED WORK

Most of the previous studies on data mining applications in various fields use the variety of data types i.e., text, image audio and video in a variety of databases. Different methods of data mining are used to extract the hidden patterns and knowledge discovery. For this purpose knowledge of the domain is very necessary. The variety of data should be collected to create the database in the specific problem domain. The next steps in this process are selection of specific data for data mining, cleaning and transformation of data, extracting patterns for knowledge generation and final interpretation of the patterns and knowledge generation described in [3].
In this paper, the estimation of this primary representation is based on a compilation of state-of-the-art research in this area, enumerated in this section. Different studies and research works have been conducted on feature selection and audio classification by employing different features and methods.

 A new feature selection algorithm FCBF is implemented and evaluated through experiments comparing with three feature selection algorithm introduced in [4]. The method also focuses on efficiency and effectiveness of feature selection in supervised learning where data content irrelevant or redundant features. George Forman [5] described a comparative study of feature selection metrics for the high dimensional domain of text classification which is focused on support vector machines and 2-class problems. It also focuses on the method for selecting one or two metrics that have the best chances of obtaining the best performance for a *dataset*. A multiclass classification strategy for the use of SVMs to solve the audio classification problem and achieve lower error rates is presented in [6]. For content based audio retrieval, it proposes a new metric, called distance-from-boundary (DFB). An SVM based approach to classification and segmentation of audio streams which achieves high accuracy described in [7]. It proposed a set of features for the representation of audio streams, including band periodicity and spectrum flux.

Various audio files, i.e., music, background sound and speeches were analyzed and classified with respect to various features extracted in a different perspective. Mainly KNN and SVM method are used for this purpose. S. Pfeiffer *et al.* [8] presented a theoretical framework and application of automatic audio content analysis using some perceptual features. Saunders [9] described a speech/music classifier based on simple features such as zero-crossing rate. Scheirer *et al.* [10] introduced features for audio classification and performed experiments with different classification models. An efficient method [11] presented for effective feature subset selection, which builds upon known strengths of the tree ensembles from large, dirty, and complex data sets (in 2009). Research work [12] to evaluate feature selection algorithms for financial credit-risk evaluation decisions and the selected features are used to develop a predictive model for financial credit-risk classification using a neural network (in 2010). S. Maldonado et al. [13] presented an embedded method that simultaneously selects relevant features during classifier construction by penalizing each feature's use in the dual formulation of SVM (in 2011). Unifying framework mentioned in [14] for feature selection based on dependence maximization between the selected features and the labels of an estimation problem, using the Hilbert-Schmidt Independence Criterion proposed in 2012. In 2013, Mauricio Schiezaro and Helio Pedrini introduces feature selection method based on the Artificial Bee Colony approach [15], that can be used in several knowledge domains through the wrapper and forward strategies and the method has been widely used for solving optimization problems. Many other works have been conducted to enhance audio classification algorithms. The basic objectives of research works are to increase the efficiency of the classification process and reduce the rate of error.

The current open issues of data mining are based on (a) development of unifying theory, (b) information network analysis, (c) process-related, biological and environmental problems, and (d) dealing with complex, non-static, unbalanced, high dimensional and cost-sensitive data. Handling of historical and real-time data simultaneously is very difficult for analytics system found in [16]. The significance of the statistical result is very important in comparing to random output. In future we have to concentrate on more research work with practical and theoretical analysis to provide new methods and technique in the field of distributed data mining. Representation of large data set and space required for storing the data these two are very important factors. In case of compression less space is required but there is no loss of information. The 2012 IDC study on Big Data [17] mentioned that in 2012, 23% of the digital universe would be useful for Big Data if tagged and analyzed. But at present only 3% of the potentially useful data is tagged, and even less is analyzed.

## 3. SYSTEM DESIGN

The Figure 1 depicted various stages of the whole process starting from database creation to classification phase. At first we have created a training database of classified and non-classified audio files. Classified audio file database is consisting of various .wav files of different music genre. In the second phase various features of an audio file are identified. Then numerical values of each feature are extracted from individual audio files for creating feature vector in the third step of system design. The next phase deals with the data analysis part. After that range of each feature for every music class is identified for further processing. In fifth step gain ratio is calculated with respect to pulse clarity to identify the splitting attribute. Three splitting attributes with maximum gain ratio is already selected and then in sixth phase, a threshold value of each feature is calculated. The next phase describes classification of audio files from testing database. In the last step we concentrate on result analysis and calculation of successful classification and rate of error.

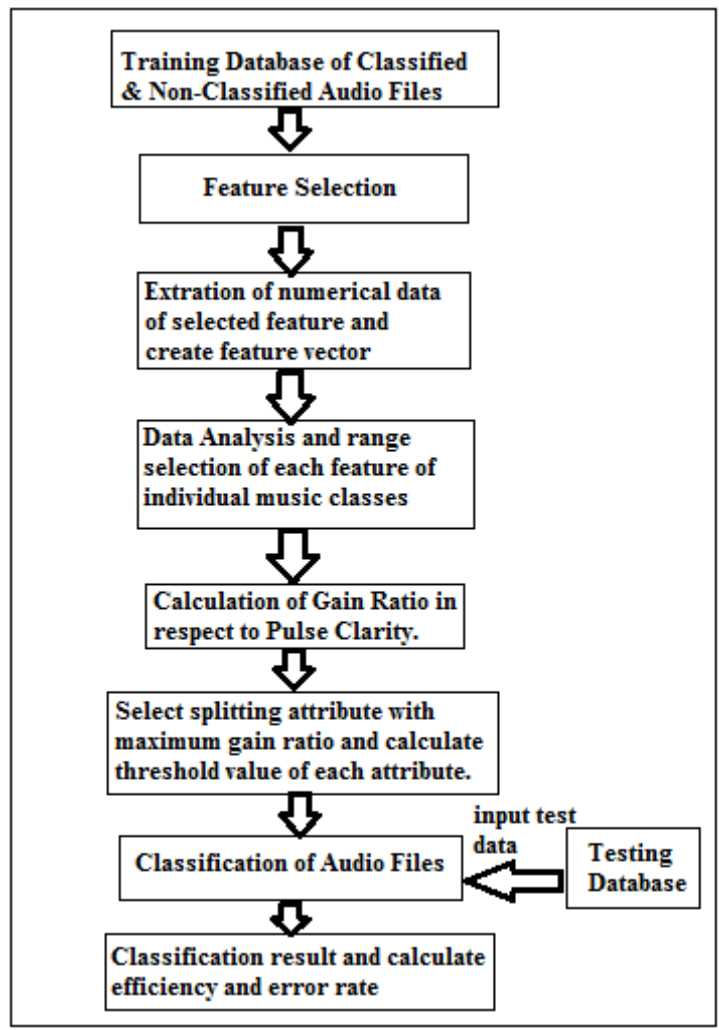

Figure 1. System Design

## 4. FEATURE SELECTION

The development of internet technology is widely increasing the use of multimedia data i.e., image, audio and video. In this paper our focus is on audio and music. Data mining techniques can be used to discover relevant similarities between music for the purpose of classifying it in a more objective manner. The backbone of most music information retrieval systems is the features extracted from audio file. The effectiveness of this recording is dependent on the ability to classify and retrieve the audio files in terms of their sound properties. Audio files basically of three types i.e. speech, music and background sound. Male and female speech files are available in .wav file format. Music files are classified into two categories - classical and non-classical music. According to genre of the track the classical music is subdivided into the chamber and orchestral. Rock, Pop, Jazz and Blues are various genres of non-classical music. Rock and Pop music is sub classified to hard rock, soft rock, techno and hip-hop music. For these work totals 11 genres of music are identified. An important phase of audio classification is feature selection. In order to obtain high accuracy for classification and segmentation, it is very important to select good features of audio files. Generally audio file analysis is based on the nature of the waveform. So the features have been selected on the basis of their numerical values. Before feature extraction, an audio signal is converted into a general format, which is .wav format.

Selected features of an audio file are sampling rate (in Hz.), temporal length (seconds/sample), rms energy, low energy, tempo (in bpm), pulse clarity, zero crossing rate (per second), roll off (in Hz.), Spectral irregularity, Pitch (in Hz.) and inharmonicity. The numerical value of each feature is computed using MIRToolBox for further analysis. RMS Energy of an audio file is represented in the Figure 2.

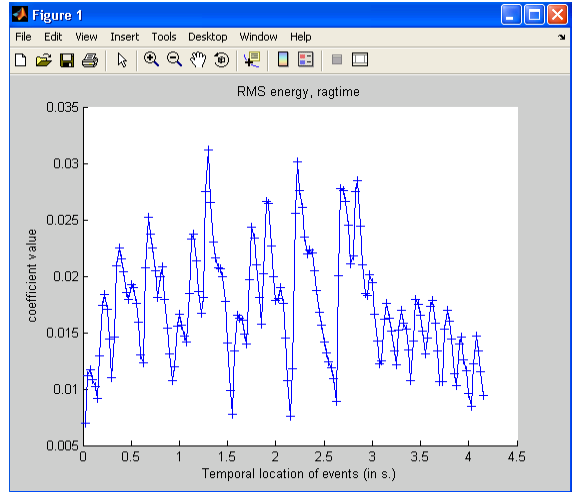

Figure 2. RMS Energy

## 5. GAIN RATIO

Information Gain and Gain Ratio are used in this work as attribute selection measures. Node N holds the tuples of partition D. The expected information needed to classify a tuple in D is as follows:

$$\text{Info}(D) = - \sum_{i=1}^{m} P_i \log_2(P_i)$$

where $P_i$ is the nonzero probability that an arbitrary tuple in D belongs to class $C_i$ and is estimated by $|C_{i,D}|/|D|$.

To partition the tuple in D on some attribute A having v distinct values $\{a_1, a_2, \ldots a_v\}$. Attribute A can be used to split D into v partitions $\{D_1, D_2, \ldots D_v\}$, where $D_j$ contains those tuples in D that have outcome $a_j$ of A. The expected information required to classify is,

$$\text{Info}_A(D) = \sum_{j=1}^{v} |D_j|/|D| \times \text{Info}(D_j)$$

where $|D_j|/|D|$ = weight of jth partition.
Information Gain is, $\text{Gain}(A) = \text{Info}(D) - \text{Info}_A(D)$
By splitting the training data set D into v partitions on attribute A Splitting Information is calculated as follows:

$$\text{SplitInfo}_A(D) = - \sum_{j=1}^{v} |D_j|/|D| \times \log_2 (|D_j|/|D|)$$

and the Gain Ratio is, $\text{GainRatio}(A) = \text{Gain}(A) / \text{SplitInfo}_A(D)$.

## 6. DATA ANALYSIS AND EXPERIMENTAL EVALUATION

The audio files of training database are already classified and the numerical values of each feature of a specific audio file are extracted using MIRToolBox. MIRToolbox is a Matlab toolbox dedicated to the extraction of musically related features from audio recordings. It has been designed in particular with the objective of enabling the computation of a large range of features from databases of audio files, which can be applied to statistical analyses described in [18]. MIRToolBox application is used with MATLAB 2012 to compute the numerical values of selected features. The files of testing database are processed and input to the application to identify the music genre of that specific audio file. The following commands are executed in a MIRToolBox application to calculate the numerical values of individual features for all files:

miraudio('b1.wav')
mirlength('b1', 'Unit', 'Second')
mirrms('ragtime')
r1 = mirrms('b1', 'Frame')
mirlowenergy(r1)
mirtempo('b1', 'Autocor')
mirpulseclarity('b1', 'MaxAutocor')
mirzerocross('b1', 'Per', 'Second')
mirrolloff('b1', 'Threshold', .85)
mirregularity('b1', 'Jensen')
mirpitch('b1', 'Autocor')
mirinharmonicity('b1', 'f0', 450.8155)

### 6.1 Non-Classified Audio File Database

Total 52 .wav files are collected for creating non-classified Database. Extracted numerical values of this feature of non-classified audio files are stored in a dataset. The data values of individual features of non-classified audio files are plotted and analyzed in a waveform and linear values are identified. Temporal Length of non-classified audio files is represented in Figure 3. Then the data values of individual features are subdivided into three groups- high, medium and low on the basis of their range of values. Pulse clarity is considered as a high-level musical dimension that conveys how easily in a given musical piece, listeners can perceive the underlying rhythmic or metrical pulsation. This Characterization of music plays an important role in musical genre recognition described in [19] which allows discrimination between genres but that differ in the degree of emergence of the main pulsation over the rhythmic texture. The notion of pulse clarity is considered in this study as a subjective measure and it is used to calculate the gain of features of all audio files. Roll Off has the highest information gain among features, so it is selected as the splitting attribute. The gain ratio of each feature is calculated for audio file analysis in respect to the feature pulse clarity. Table 1 consists of gain ratio of each feature in respect of various music classes. The Roll Off feature with the maximum gain ratio is selected as the splitting attribute in case of non-classified audio files.

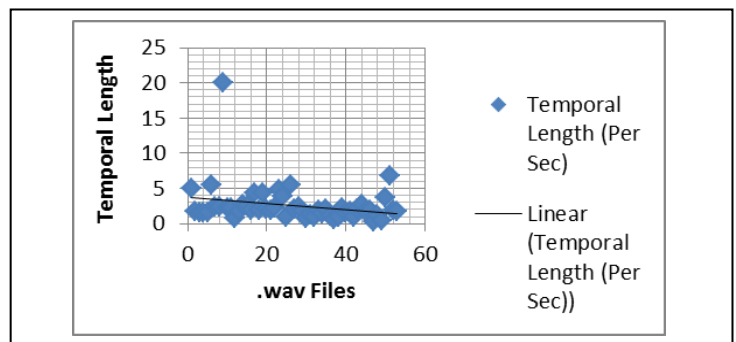

Figure 3. Temporal Length

Table 1. Gain_Ratio of Non-Classified Audio Files

| Feature | Gain | Gain Ratio |
|---|---|---|
| Sampling Rate | .2923 | .164 |
| Temporal Length | .0048 | .00687 |
| RMS Energy | .0412 | .0513 |
| Low energy | .0214 | .0198 |
| Tempo | .1022 | .1228 |
| Zero Crossing Rate | .1562 | .0996 |
| Roll Off | .3037 | .2428 |
| Spectral Irregularity | .1392 | .0981 |
| Pitch | .0412 | .0262 |
| Inharmonicity | .0445 | .1159 |

## 6.2 Classified Audio File Database

The Classified Database is consisting of 110 audio files which is based on some basic classes i.e., Blues, Classical, Chamber, Orchestral, Jazz, Pop, Hip Hop, Techno, Rock, Hard rock and Soft rock music. 10 .mp3 files for each class are collected for the creation of the database. Then the files are converted into .wav format for processing. The sampling rate of all files is fixed to 44100 Hz. These extracted values are saved in a database for further analysis. The data values of individual features are plotted and analysed for all classes. From the graph layout we can get the minimum and maximum range of a specific feature for a particular music class. Figure 4 shows the data ranges of various features. Based on different ranges the data sets are divided into three groups – low, medium and high.

| Types | Temporal Length | RMS Energy | Low Energy | Tempo(bpm) | Pulse Clarity | Zero crossing rate | Rolloff(Hz) | ectral Irregular | Pitch(Hz) | Inharmonicity |
|---|---|---|---|---|---|---|---|---|---|---|
| Blues | 67.030-218.044 | .039-.066 | .473-.601 | 104.547-167.276 | .021-.594 | 721.237-1900.420 | 2522.161-7131.948 | .005-.641 | 110.631-986.892 | .468-.515 |
| Classical | 62.537-200.112 | .034-.057 | .504-.637 | 70.926-160.200 | .030-.547 | 470.025-837.309 | 1118.381-5676.943 | .181-.835 | 83.007-355.879 | .401-.482 |
| Chamber | 133.251-223.622 | .037-.073 | .466-.629 | 91.508-172.097 | .041-.653 | 583.278-2395.851 | 1846.220-6689.256 | .141-.543 | 154.685-506.055 | .419-.496 |
| Orchestral | 63.817-183.249 | .046-.07 | .506-.592 | 106.047-190.455 | .060-.861 | 443.363-1699.553 | 2318.941-5467.835 | .176-.985 | 195.730-853.233 | .458-.550 |
| Jazz | 125.350-236.266 | .050-.069 | .454-.645 | 96.695-183.722 | .150-.691 | 613.763-1894.917 | 1804.752-5890.593 | .071-.882 | 116.615-1017.057 | .469-.508 |
| Pop | 63.387-221.780 | .043-.078 | .484-.582 | 99.969-182.133 | .141-.736 | 515.761-2735.099 | 3571.568-6644.087 | .049-.769 | 157.633-945.820 | .400-.514 |
| HipHop | 99.892-221.951 | .039-.066 | .465-.572 | 77.101-193.077 | .237-.672 | 1015.925-3284.513 | 5211.119-8124.579 | .075-.890 | 114.463-1006.787 | .476-.503 |
| Techno | 80.106-220.435 | .039-.095 | .462-.672 | 114.681-189.943 | .529-.878 | 816.616-2548.420 | 3221.569-8289.526 | .094-.609 | 118.570-1067.258 | .436-.524 |
| Rock | 113.293-236.931 | .055-.070 | .468-.568 | 87.217-157.269 | .123-.521 | 1278.300-3066.453 | 3898.015-6736.528 | .081-.395 | 78.623-903.383 | .474-.508 |
| Hard Rock | 104.635-224.484 | .055-.070 | .494-.564 | 100.221-189.410 | .274-.643 | 929.503-2250.428 | 4023.429-6179.104 | .110-.307 | 116.642-1080.453 | .473-.507 |
| Soft Rock | 47.479-205.546 | .057-.086 | .492-.570 | 104.660-176.061 | .082-0.676 | 1122.605-1994.774 | 3626.326-7124.883 | .092-.436 | 198.253-1032.027 | .473-.508 |

Fig4. Feature_DataRange

Figure 4. Feature_DataRange

After that information gain and gain ratio are calculated in respect to pulse clarity of each feature corresponding to a particular music class. Maximum and average values of gain ratio are calculated as $Max_{Gain}$ and $Avg_{Gain}$ respectively for individual feature. Then threshold value is calculated from the average value of $Max_{Gain}$ and $Avg_{Gain}$. Figure 5 Shows that, Roll off, Zero

Crossing rate and Tempo have maximum threshold value. So these three features are selected as the main criterion to classify an audio file.

| Gain Ratio in respect to Pulse Clarity | | | | | | | | | |
|---|---|---|---|---|---|---|---|---|---|
| Music Type | Temporal Length | RMS Energy | Low Energy | Tempo | ZeroCrossing Rate | RollOff | Spectral Irregularity | Pitch | Inharmonicity |
| Blues | 0.5236 | 0.3077 | 0.5368 | 0.0982 | 0.3244 | 0.7073 | 0.0876 | 0.2258 | 0.2437 |
| Classical | 0.3012 | 0.2367 | 0.3508 | 0.4343 | 0.3997 | 0.1092 | 0.1379 | 0.2097 | 0.1413 |
| Chamber | 0.3044 | 0.2347 | 0.5459 | 0.3045 | 0.3797 | 0.3567 | 0.0876 | 0.3567 | 0.2437 |
| Orchestral | 0.3897 | 0.2652 | 0.2157 | 0.1614 | 0.2897 | 0.3168 | 0.1414 | 0.1092 | 0.4223 |
| Jazz | 0.4654 | 0.4039 | 0.4375 | 0.4999 | 0.2876 | 0.3189 | 0.3428 | 0.5324 | 0.153 |
| Pop | 0.1317 | 0.4849 | 0.3101 | 0.4462 | 0.5698 | 0.2024 | 0.3543 | 0.0505 | 0.4126 |
| HipHop | 0.4106 | 0.1605 | 0.1543 | 0.5591 | 0.3543 | 0.0905 | 0.2252 | 0.446 | 0.1543 |
| Techno | 0.5146 | 0.3242 | 0.5146 | 0.1127 | 0.362 | 0.3972 | 0.3927 | 0.2252 | 0.2076 |
| Rock | 0.3508 | 0.2437 | 0.14 | 0.2432 | 0.3242 | 0.2098 | 0.1123 | 0.4265 | 0.3598 |
| Hard Rock | 0.362 | 0.1127 | 0.2933 | 0.5519 | 0.4276 | 0.1054 | 0.1123 | 0.0984 | 0.3466 |
| Soft Rock | 0.3925 | 0.3492 | 0.2863 | 0.3762 | 0.3429 | 0.3268 | 0.0801 | 0.1011 | 0.266 |
| Max | 0.5236 | 0.4849 | 0.5459 | 0.5591 | 0.5698 | 0.7073 | 0.3927 | 0.5324 | 0.4223 |
| Avg | 0.3770 | 0.2839 | 0.3441 | 0.3443 | 0.3693 | 0.2855 | 0.1886 | 0.2529 | 0.2683 |
| Threshold | 0.4503 | 0.3844 | 0.4450 | 0.4517 | 0.4695 | 0.4964 | 0.2906 | 0.3926 | 0.3453 |

Figure 5. GainRatio_PulseClarity

We've developed an application in PHP to implement the different functionalities i.e., feature extraction and classification. Training databases of classified and non-classified audio files and testing database are also created using MySQL. At first, an audio file from testing database is entered into the application for feature extraction and the numerical values of each feature of the audio file are displayed. From this data value system can identify the class of that music file. In this method, three splitting attributes which have a maximum gain ratio can be easily calculated from the gain_ratio table. For classification phase, the values of Roll of, Zero Crossing rate and Tempo are fixed to threshold value which is already calculated and selected as the basic criterion of classification. The threshold value of each feature is calculated as follows:

$$\text{Threshold}_{Fi} = [\{(\text{Max}_{Fi} + \text{Avg}_{Fi})/2\} + \{(\text{Max}_{Fi} - \text{Avg}_{Fi})/4\}]$$

During the classification phase of an audio file (from testing database) the system compares the values with threshold values. If the values are less than or equal to threshold value then it is successfully classified and music class is identified. Out of 110 music files 9 files are not classified as the value exceeds the threshold limit.

| Total | Successful (91.82%) | Unsuccessful (8.18%) | | | | | |
|---|---|---|---|---|---|---|---|
| | | Jazz | Orchestral | Pop | Hip Hop | Techno | Rock | Hard Rock |
| 110 | 101 | 2 | 1 | 1 | 1 | 2 | 1 | 1 |

Figure 6. Classification Result

Figure 6 represents the classification result on testing database. All test music files of classes Blues, Classical, Chamber and Soft rock are classified successfully and no error occurs. For improvement of efficiency of classification and to get the optimal solution threshold value plays an important role. 91.82% files of testing database are correctly classified and the error rate of unsuccessful classification is 8.18%. So, our approach is effective with respect to music genre classification accuracy in [20]. By dimensionality reduction top three features have been selected which are used for classification. The classification result is not affected by this method and we can improve the percentage of successful classification above 90%, which can further improve in the future on the basis of performance of the system and the result of success and error in classification.

## CONCLUSIONS

This approach includes the method to consistently and precisely identify the features that take part in the classification of an audio file. We have described in detail an audio classification scheme that uses Gain Ratio to select splitting attribute for classification into various music genres. This research work presented here is based on feature selection, extraction, analysis of data, feature selection using gain ratio and finally a classification stage using splitting criterion. For calculation of gain ratio we have included Pulse Clarity feature which has high discrimination power among different features and it ensures that the system can achieve high accuracy. All audio files are of .wav format and the sampling rates are fixed. The audio files belonging to any particular music genre share a similarity in a range of values of their features and hence make it possible to discover the pattern and then classify the audio file accordingly. This work emphasizes both the theoretical concept as well as gives insight into the practical application program. Experimental results showed that the scheme is very effective and the total accuracy rate is over 90%. There are many interesting directions that can be explored in the future. To achieve this, we need to concentrate on the selection of more audio features that can be used to characterize the audio content. In the future, our audio classification scheme will be improved to discriminate more audio classes, speeches, background music or any other sound. We will also focus on developing an effective scheme to apply data mining techniques to improve the efficiency of the classification process. The future scope of the work is to improve the quality of the audio file by improving the quality of sound by reducing the noise.

**Authors**

**Jayita Mitra**

Presently working as Assistant Professor in Dept. of Information Technology at Camellia Institute of Technology, Kolkata. She has 6.5 Years of academic experience and currently pursuing her research in Data Mining from Jadavpur University.

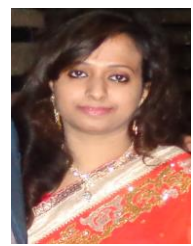

**Diganta Saha**

Presently working as Associate Professor in Dept. of Computer Science and Engineering at Jadavpur University. His area of specialization are Machine Translation, Natural Language processing, Mobile Computing and Pattern Classification.

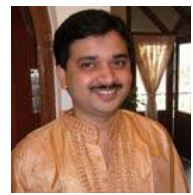